\documentclass[letterpaper, 10pt, journal, twoside]{IEEEtran}

\usepackage{hyperref}
\usepackage[latin9]{inputenc}
\usepackage{amsmath}
\usepackage{amssymb}
\usepackage{graphicx}
\usepackage{float}

\usepackage{enumerate}
\usepackage{tikz}
\usetikzlibrary{shapes,arrows,automata,calc,trees,positioning,fit,shapes,calc,patterns,3d}
\usepackage{pgfplots}
\usepgfplotslibrary{fillbetween}
\usepackage{booktabs}
\usepackage{xparse}
\usepackage{subfigure}

\usepackage{algorithm}
\usepackage{algpseudocode}

\usepackage[normalem]{ulem}
\begin{document}

\title{ROS-NetSim: A Framework for the Integration of \\ Robotic and Network Simulators}

\author{Miguel Calvo-Fullana$^{1}$, Daniel Mox$^{2}$, Alexander Pyattaev$^{3}$, \\Jonathan Fink$^{4}$, Vijay Kumar$^{2}$ and Alejandro Ribeiro$^{2}$%
\thanks{This work was supported by ARL DCIST CRA W911NF-17-2-0181 and the Intel Science and Technology Center for Wireless Autonomous Systems.}
\thanks{$^{1}$Miguel Calvo-Fullana was with the University of Pennsylvania, Philadelphia, PA, USA. He is now with the Massachusetts Institute of Technology, Cambridge, MA, USA.
{\tt\footnotesize cfullana@mit.edu}}%
\thanks{$^{2}$Daniel Mox, Vijay Kumar and Alejandro Ribeiro are with the University of Pennsylvania, Philadelphia, PA, USA.
{\tt\footnotesize \{mox,kumar,aribeiro\}@seas.upenn.edu}}%
\thanks{$^{3}$Alexander Pyattaev is with Tampere University, Tampere, Finland.
{\tt\footnotesize alexander.pyattaev@tuni.fi}}%
\thanks{$^{4}$Jonathan Fink is is with the U.S. Army Research Laboratory, Adelphi, MD, USA.
{\tt\footnotesize jonathan.r.fink3.civ@mail.mil}}%
}


\maketitle

\begin{abstract}
Multi-agent systems play an important role in modern robotics. Due to the nature of these systems, coordination among agents via communication is frequently necessary. Indeed, Perception-Action-Communication (PAC) loops, or Perception-Action loops closed over a communication channel, are a critical component of multi-robot systems. However, we lack appropriate tools for simulating PAC loops. To that end, in this paper, we introduce \emph{ROS-NetSim}, a ROS package that acts as an interface between robotic and network simulators. With ROS-NetSim, we can attain high-fidelity representations of both robotic and network interactions by accurately simulating the PAC loop. Our proposed approach is lightweight, modular and adaptive. Furthermore, it can be used with many available network and physics simulators by making use of our proposed interface. In summary, ROS-NetSim is (i) Transparent to the ROS target application, (ii) Agnostic to the specific network and physics simulator being used, and (iii) Tunable in fidelity and complexity. As part of our contribution, we have made available an open-source implementation of ROS-NetSim to the community.
\end{abstract}


\section{Introduction}
\label{sec:introduction}

\IEEEPARstart{W}{ireless} communications and wireless networks are ubiquitous in the modern world. With the advent of disruptive technologies such as autonomous driving \cite{campbell2010autonomous,geiger2012we} and the Internet of Things \cite{atzori2010internet,zanella2014internet}, the presence of wireless communications will only grow stronger. As a unifying factor, all of these new technologies are characterized by some degree of autonomy. This is a problem that is at the core of robotics. Robotic systems make use of sensors to reason about the environment and consequently take actions, forming what is known as the Perception-Action (PA) loop. In many systems, specially those that are multi-agent in nature, this Perception-Action loop is closed over a communication channel, giving rise to the aptly named Perception-Action-Communication (PAC) loop \cite{yang2018grand}. An important question then arises: How much fidelity does one need in the communication component of the PAC loop in order to accurately simulate the behavior of the aggregate robotic system?

The answer to the previous question is necessarily going to depend on many aspects. However, one thing is clear, we must be able to simulate the PAC loop, which is a form of feedback closed over a communication channel. Thus, if the communication aspects of the loop are not accurately represented, undesired behavior can be introduced into the action component. This in turn can cause a cascading effect, resulting in behaviors that do not accurately represent reality. In effect, this can be observed in practice, as severe deviations can be frequently found between simple simulation and experimental deployment of systems with PAC loops \cite{fink2011robust,stephan2017concurrent,mox2020mobile}.

In this sense, PAC loops have been considered a key component of multi-robot systems \cite{yang2018grand}. However, current solutions to accurately simulate them are severely limited. On one hand, popular and well developed network simulators are widely available and used by the research community. Examples of which are ns-3 \cite{riley2010ns} and OMNeT++, among many others. On the other hand, celebrated software, like the Robot Operating System (ROS) \cite{quigley2009ros} powers a large part of robotics research. These components are further enhanced by the use of tools capable of highly detailed physics simulation and realistic rendering, such as Gazebo \cite{koenig2004design} and AirSim \cite{airsim2017fsr}, among others. However, these are two distinct and separate areas of research and development, network simulators attempt to represent the traffic over a communication network, while robotic simulators focus on the physical interactions of robots with their environment. By design, these simulators are unaware of the other component, limiting their ability of simulating PAC loops. As a naive approach, one could always collect data from the wireless channel and then introduce it into the robotic simulator of choice. This is a traditional approach taken by the community \cite{hsieh2008maintaining, fink2009experimental, balaguer2012combining, howard2003experimental}. However, this approach has a major drawback. In the presence of a PAC loop, the underlying control loop, which defines the behavior of the robotic agents, mandates that the simulation be done jointly, or else, suffer the consequences of simulated behavior ill-representing reality. 

\subsection{Related Work}

To overcome the previous issues, some approaches have been taken to integrate network simulators into physical system simulators in an approach known in general as co-simulation \cite{li2014co,gomes2018co}. In recent times, due to their popularity, most interest in robotic and network co-simulation has focused on Unmanned Aerial Vehicles (UAVs) \cite{acharya2020cornet,zema2017cuscus,marconato2017avens,baidya2018flynetsim}. These approaches, while valid in their respective UAV-centric applications, suffer from several downsides. Broadly speaking, (i) they are designed with UAV applications in mind and do not necessarily translate well to other robotic platforms; (ii) they do not provide a cross-simulator interface, as they are tightly integrated with their respective choice of physics and network simulator; (iii) the simulation of communication only accounts for agent positions and not the geometry of the surrounding environment; (iv) the network traffic from the target application is not captured for simulation, instead network traffic is independently generated by agents at positions given by the physics simulator; (v) they do not always have complete synchronism across simulators, either being unsynchronized, or only synchronized in one direction of simulation (able to stop only one of the simulators).

When dealing with general purpose robotic systems, the most remarkable example of joint robotic and network simulation is RoboNetSim \cite{kudelski2013robonetsim}, which expands on the ARGoS robotic simulator \cite{pinciroli2012argos} by providing integration and synchronization with the ns-2/ns-3 network simulator. Compared with the previous approaches, RoboNetSim is designed with general purpose robotics in mind, not only UAVs, and it provides a synchronization mechanism capable of bridging the discrete-time and discrete-event nature of physics and network simulators, respectively. However, it still suffers from some of the previously mentioned downsides. Mainly, RoboNetSim is tightly integrated with the ARGoS and ns-3 simulators, lacks a packet capture mechanism and does not extract channel information from the geometry of the environment. This results in the limited portability of target applications, as they need to be explicitly implemented on the RoboNetSim simulation platform. Unavoidably, the issues still outstanding mean that there is no portable, general-purpose solution when one desires to simulate multi-agent systems with realistic control loop interactions and communications. Something that we aim to address in this work.

\subsection{Contribution}

We introduce \emph{ROS-NetSim}, our integrated approach for joint robotic and network simulation. Our main intent is to leverage the powerful and widely used resources currently available, as such, we do not intend to introduce a new simulator. Instead, we propose to integrate current robotic and network simulators via the use of a carefully designed interface. We emphasize the following three distinct elements. (i) The system is transparent to the ROS application with only configuration information needed to be specified; (ii) The system is agnostic to the choice of both the network and the physics simulator; (iii) The simulation description is tunable to account for a large range of communication fidelity and complexity. The proposed interface consists of two main blocks, a physical coordinator and a network coordinator. The former, extracts from the physics simulator a geometrical representation of the communication channels among agents that is shared with the network simulator via the network coordinator. The latter, the network coordinator, captures the traffic generated by the ROS nodes and forwards it to the network simulator for accurate processing. Furthermore, all the systems maintain the same clock timing via the sharing of synchronization messages. As a companion contribution to this paper, we have released a ROS package, ROS-NetSim, which implements our proposed simulation architecture\footnote{\url{https://github.com/alelab-upenn/ros-net-sim}.}. 

\section{The ROS-NetSim Simulation Architecture}
\label{sec:architecture}

\begin{figure}[t]
    \centering
    \includegraphics[scale=1]{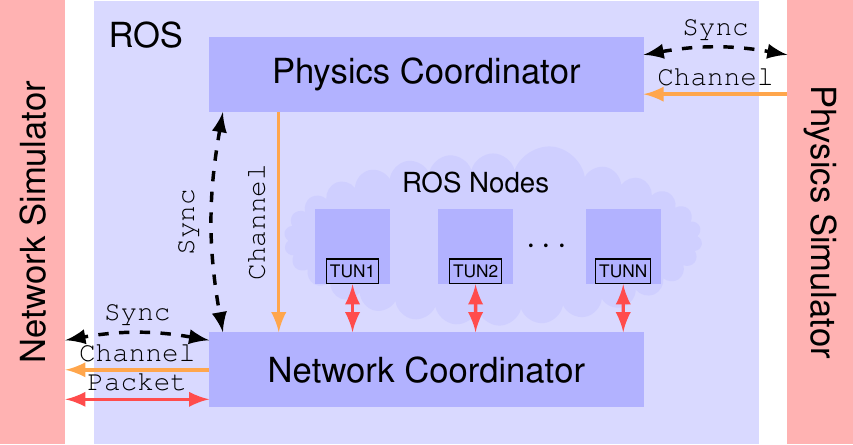} 
    \caption{The ROS-NetSim system architecture.}
    \label{fig:simulator_architecture}
\end{figure}

As mentioned previously, the main contribution of this work is the introduction of ROS-NetSim, an architecture which allows for the concurrent operation of robotic and network simulators. To this end, ROS-NetSim overcomes the previously identified issues in order to attain the general purpose simulation of networked robots. Joint simulation of the physics and network components is required for the accurate representation of these systems. This requires the close interaction between different simulators, which at their core, operate in different manners. Namely, physics simulator are time-based simulators, while network simulators are event-based. ROS-NetSim addresses these differences by using a system-wide synchronization window, based on the exchange of state messages, attaining complete synchronism during the joint simulation. Furthermore, ROS-NetSim addresses several outstanding issues found in previous joint simulation systems. Among them is the effect of the surrounding physical environment on the wireless communication performance. ROS-NetSim uses a modular channel abstraction which is queried to the physics simulator at each synchronization window and provided to the network simulator, resulting in the ability to produce highly detailed simulations of communication conditions given the state of the robotic agents and their environment. Finally, ROS-NetSim addresses the issue of transparency to the target application, by introducing a packet capture mechanism based on network tunnels. This allows for target ROS code to be embedded into ROS-NetSim without the need to explicitly design for it. Thus, network traffic is seamlessly captured by ROS-NetSim and forwarded to the network simulator for further processing. An overview of the resulting architecture is illustrated in Figure \ref{fig:simulator_architecture}.

The system is composed of several blocks. As a main component, the ROS-NetSim system is built on top of the Robot Operating System (ROS) \cite{quigley2009ros} framework. Two units rest outside of ROS, the physics simulator and the network simulator. These components can be anything that the user of ROS-NetSim desires, as long as certain coordination and message passing mechanisms are implemented. Standard choices would be, for example, ns-3 \cite{riley2010ns} as a network simulator and Gazebo \cite{koenig2004design} for a physics simulator. These external entities exchange information with each other via ROS-NetSim, specifically the two coordination units (one handling the physics aspects and the other the network ones). Among other tasks, these coordinators maintain timing (clock) synchronicity between the external simulators and the ROS nodes. In particular, the physics coordinator extracts from the physics simulator a geometrical, material representation of the communication channel between each pair of agents in the simulation, which is passed on to the network coordinator which forwards this information to the network simulator. Additionally, the network coordinator captures all the network traffic generated by the ROS nodes, sharing it with the network simulator for appropriate characterization. In this way, ROS nodes are launched inside this environment and generate real network traffic subject to the underlying physics-informed network simulation.

\subsection{ROS Space}\label{sec:ros_space}

One of the main design choices of the ROS-NetSim system is transparency towards the ROS target application. In other terms, this means that the rest of the simulation (not ROS-NetSim) is launched as usual. The ROS nodes corresponding to the agents executing communication tasks do not need to be modified to make use of our system architecture. These nodes handle communication in the usual way, as they would do in an actual experiment. They use network sockets and IP addresses to transfer information across agents. 

In order to accomplish this transparency, ROS-NetSim must be able to correlate the agent identifier used by the physics simulator with the IP address tied to its network interface (or multiple addresses if the node uses multiple interfaces). The physics coordinator uses the agent identifier to collect agent state and environment information from the physics simulator and the network coordinator uses the IP addresses of the network interfaces to capture and release network traffic. Beyond these two configuration parameters, the nodes launched in ROS space are unaware of the presence of ROS-NetSim and the simulation of their data traffic. A detailed treatment of the physics and network coordinators follows.

\section{Physics Simulation and Coordination}
\label{sec:phy_sim_and_coord}

Accurate simulation of physics and dynamics is of paramount importance in robotics. Modern platforms boast astonishing levels of realism by capturing the minute details of each asset. Our system leverages this fidelity in order to infer how the environment affects each communication channel. If we reduce the communication channel between a pair of agents to its bare physical representation, it mainly depends on the position and orientation of each agent and the surrounding environment. Following this principle, we use the physics simulator to generate a description of the communication channel that captures both geometric and material information. This description, coupled with an accurate representation of the communication technology being used (provided by the network simulator) results in a realistic communication channel representation. The following sections detail how the physics coordinator in our system handles extracting this information from the physics simulator and passing it on to the network simulator in a synchronized manner.

\subsection{Channel Abstraction}
\label{sec:channel_abstraction}

\begin{figure}[t]
    \centering
    \includegraphics[scale=1]{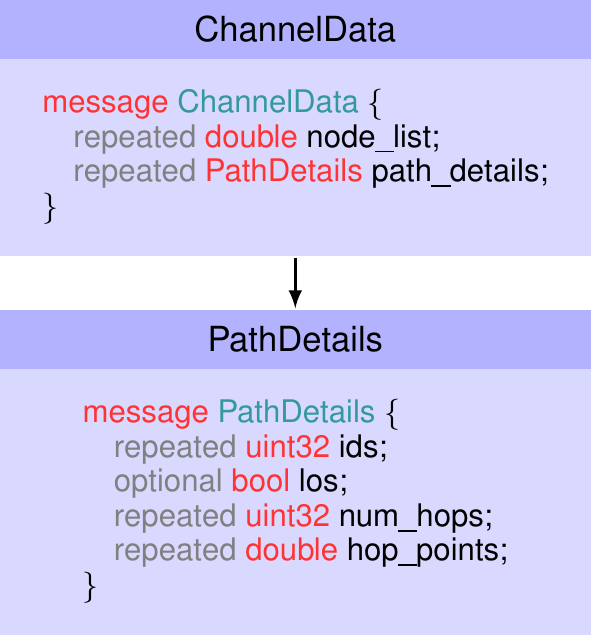} 
    \caption{The \texttt{ChannelData} message.}
    \label{fig:msgChannelData}
\end{figure}

We encode the channel abstraction in the form of a \texttt{ChannelData} protobuf message \cite{protobuf}, as illustrated in Fig. \ref{fig:msgChannelData}. This message is packed with information extracted from the physics simulator by the physics coordinator and eventually forwarded to the network simulator as described later in Section \ref{sec:physics_coordinator}. In our provided codebase, we have implemented these hooks for Gazebo, our physics simulator of choice, but this channel extraction routine can be implemented for other physics simulators as well.

Each \texttt{ChannelData} message contains channel information about all the agents in the simulation. Following standard protobuf terminology, each message consists of two repeated fields \texttt{node\_list} and \texttt{path\_details}. The \texttt{node\_list} field corresponds to a concatenation of $\mathbb{R}^7$ vectors containing the $[x,y,z]$ positions and $[x,y,z,w]$ orientation quaternion of each agent in free space. The field \texttt{path\_details} is packed with \texttt{PathDetails} messages which capture the various paths a signal may travel between two nodes. In \texttt{PathDetails}, the field \texttt{ids} stores the two nodes trying to communicate and \texttt{los} is true for line-of-sight and false if not. In either case, a signal may travel multiple paths through the environment enroute to its destination. The number of hops in each path is stored in the \texttt{num\_hops} array and the corresponding positions in the environment along with the incurred transition loss are stored as $[x,y,z,l]$ tuples in the \texttt{hop\_points} array. Note that different environmental interactions (reflection, penetration, etc.) and material properties can be represented by setting $l$ appropriately.

\begin{figure}[t]
    \centering
    \subfigure[Disk models.]{
    \includegraphics[scale=1]{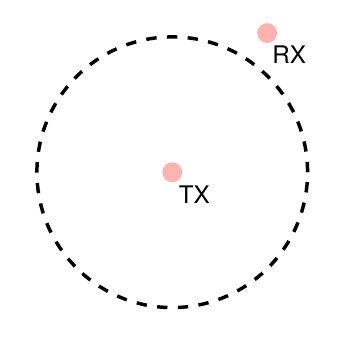}     
	 }
    \subfigure[Statistical fading models.]{
    \includegraphics[scale=1]{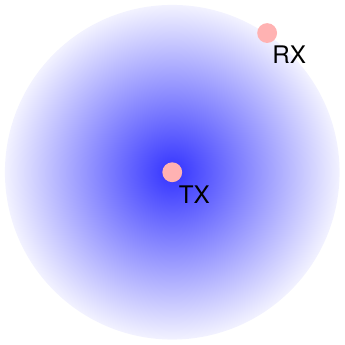}     
	 }\\
    \subfigure[LOS/NLOS models.]{
    \includegraphics[scale=1]{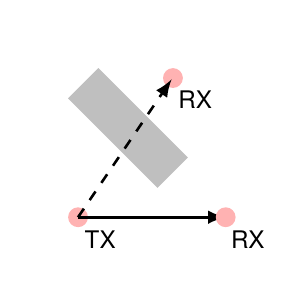}     
	 }    
    \subfigure[Raytracing models.]{
    \includegraphics[scale=1]{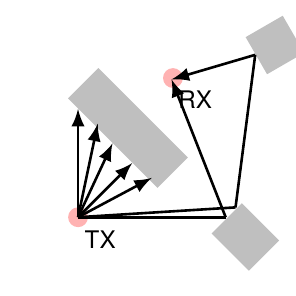}     
	 }    	 	 
    \caption{Different possible channel models that can be reproduced by a network simulator using our proposed channel abstraction. They are ordered from left to right in complexity. The upper row represents possible levels of complexity for each individual path ranging from simple on/off links (disk models) to more complex statistical representations (fading models). The bottom row shows the complexity in the number of paths that the wireless signal is considered to travel. They can range from simple single-hop LOS/NLOS paths to multi-hop and raytracing models.} 
    \label{fig:fidelity}
\end{figure}

The modularity of the \texttt{PathDetails} message can be used to cover a wide range of communication channel abstractions ranging from simple disk models to LOS/NLOS models, multi-path models and even raytracing models, as shown in Figure \ref{fig:fidelity}. The choice of which channel model to use is left to the user and their specific application needs, accuracy/complexity requirements and physics and network simulator capabilities. The Gazebo implementation that we provide extracts LOS/NLOS information.

\subsection{Synchronization}\label{sec:physics_coordinator}

In addition to gathering channel information, the physics coordinator also maintains synchronicity between the physics simulator and network coordinator (which in turn controls the network simulator described later in Section \ref{sec:network_coordinator}). More specifically, the physics coordinator exerts control over the physics simulation by starting and stopping physics updates (i.e., stepping the simulation a certain number of fixed timesteps) at will. This allows both the physics simulator and the network simulator via the network coordinator to be run based on the same clock and for the physics and network coordinators to exchange vital state information at regular timesteps. This requirement is fundamental to our system and is described in more detail in Section \ref{sec:time_synchronization}.

\begin{figure}[t]
    \center
    \includegraphics[scale=1]{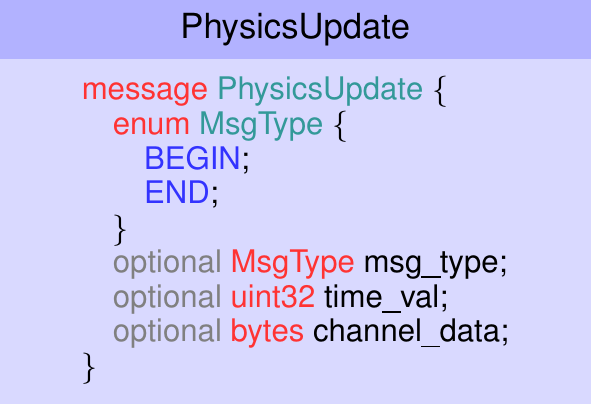}         
    \caption{The \texttt{PhysicsUpdate} message.}
    \label{fig:msgPhysicsUpdate}
\end{figure}

The physics coordinator passes information to the network coordinator in the \texttt{PhysicsUpdate} protobuf message shown in \ref{fig:msgPhysicsUpdate}. This message is composed of a \texttt{msg\_type} field, indicating whether the simulation is beginning or ending for a specific interval, a \texttt{time\_val} field indicating the timestamp of the interval being simulated, and a \texttt{channel\_data} field storing a compressed version of the \texttt{ChannelData} message described previously. In this way, the network coordinator receives all the information necessary to update the network simulator according to the state of the robots in the physics simulator.

\section{Network Simulation and Coordination}
\label{sec:net_sim_and_coord}

The network simulator aims to replicate the complicated, noisy conditions present in real world networked systems. It simulates the exchange of information between agents over a network by taking into account a variety of factors including: transport layer communication, such as unreliable transport (via UDP) or reliable transport (via TCP); routing protocols, which decide where packets should be sent enroute to their destination; and the effect of finite receiver/transmission buffers, which can backup and overflow. Note that these components vary widely across communication technologies. For example, Wi-Fi, Zigbee and Bluetooth all operate at different frequencies, support different bandwidths, execute different communication protocols, and enforce different channel access mechanisms. The network simulator allows us to accurately represent these intricacies that operate in concert and ultimately affect transmission rates and packet delay experienced by robotic agents relying on communication to close critical PAC loops. Just as with the physics simulator, the required fidelity of the network simulator varies depending on the application and it is left to the user to set appropriately.

The following sections detail our approach to capturing network traffic and provide an overview of the network coordinator, which acts as the interface between the network simulator and the physics coordinator.

\subsection{Processing of Data Traffic}
\label{sec:data_traffic}

In order to interact with the network simulator, we control traffic at the standard OSI layer 3, corresponding to IP traffic. This is accomplished using TUN virtual network interfaces, which we use to capture the layer 3 traffic of the nodes in the simulation. A TUN interface is created for each of the IP addresses listed in the configuration file (see Section \ref{sec:ros_space}). As per usual operation, ROS nodes bind sockets to these addresses and communicate over them. However, unknowingly  to them, this traffic is redirected via the TUN interfaces to the network coordinator which forwards the information of these packets to the network simulator. One useful feature of these virtual interfaces is that they possess many of the same properties as standard network interfaces. Thus the state of the TUN interface, up and down status, subnet, signal strength and so on, can also be simulated; in this way, algorithms making use of information such as RSSI are able to do so in a completely transparent manner.

\subsection{Network Coordinator}\label{sec:network_coordinator}

\begin{figure}[t]
    \centering
    \includegraphics[scale=1]{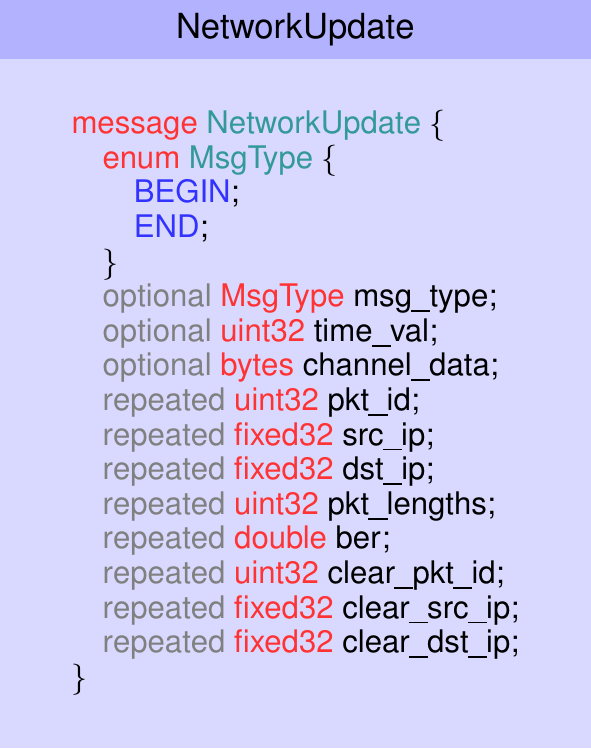}         
    \caption{The \texttt{NetworkUpdate} message.}
    \label{fig:msgNetworkUpdate}
\end{figure}

As mentioned before, the network coordinator interfaces with the physics coordinator and controls the network simulator, performing two main functions. First, the network coordinator takes the channel representation stored in the \texttt{PhysicsUpdate} received from the physics coordinator and applies it to the network simulator. This ensures the state of the network operated on by the network simulator remains consistent with the state of the agents in the physics simulator. Second, the network coordinator controls traffic on the TUN interfaces used for communication between different agents in ROS. When a packet is received on one of the interfaces, the network coordinator stores the packet data in a queue, creates a unique id (\texttt{pkt\_id}), and extracts packet length (\texttt{pkt\_lengths}), source IP (\texttt{src\_ip}), and destination IP (\texttt{dst\_ip}) information. This information gets stored in the corresponding fields of the \texttt{NetworkUpdate} message shown in Fig. \ref{fig:msgNetworkUpdate}, which gathers information about all the packets received during a simulation timestep. At the next simulation update this \texttt{NetworkUpdate} message is passed to the network simulator which advances its simulation and updates the fields \texttt{clear\_pkt\_id}, \texttt{clear\_src\_ip}, and \texttt{clear\_dst\_ip} of the packets that should be released with a specific bit error rate \texttt{ber} that should be applied. In this way, packet delay is simulated up to the granularity of the synchronization window. Finally, the network coordinator applies the bit error rate to the specific packets and releases them into the corresponding virtual network interface, allowing the ROS node to receive it. At the end of this process, the ROS node receives a truly simulated packet with realistic bit error rates.

\section{Message Passing and Synchronization}\label{sec:time_synchronization}

Due to the different design principles behind physics and network simulators, synchronization between them needs to be carefully planned. By design, physics simulators are time-based, they keep track of time and simulation evolves at a given time step. On the other hand, network simulators are event-driven. The state of the network simulator switches as events occur, without maintaining a constant time step between events. In order to synchronize between the two different approaches of the simulation we use a sliding window mechanism. Using this mechanism, we capture and track network events over the window period and allow the network simulator to step up to the end of the window.

Furthermore, given the many choices of simulators and model complexity, the physics and network simulators are not expected to run at the same speed or in real time. Thus, to function properly, all the components of our architecture need to operate in a synchronized fashion. In other words, they need to run using the same clock. This goal is accomplished following the synchronization protocol in Algorithm \ref{alg:sync_alg}. To begin, the physics and network simulators are advanced a fixed simulation window size $W$ triggered by receipt of the respective \texttt{PhysicsUpdate} and \texttt{NetworkUpdate} messages with the \texttt{msg\_type} populated with \texttt{BEGIN}. When the physics simulator has advanced the simulation by a window $W$ it returns the \texttt{PhysicsUpdate} message populated as described in Section \ref{sec:phy_sim_and_coord} and with the \texttt{msg\_type} field set to \texttt{END}. Likewise, the network simulator finishes processing the network update over the window $W$ and returns the \texttt{NetworkUpdate} message populated as described in Section \ref{sec:net_sim_and_coord} also with the \texttt{msg\_type} field set to \texttt{END}. In the next step, the physics coordinator passes its \texttt{NetworkUpdate} message to the network coordinator with the \texttt{msg\_type} field set to \texttt{BEGIN} and the network coordinator sends the an empty \texttt{NetworkUpdate} message to the physics coordinator with the \texttt{msg\_type} field set to \texttt{BEGIN} and the process starts over. Since both the network coordinator and the physics coordinator must wait for the other to complete their simulation update before proceeding, the physics simulator and the network simulator remain in sync over the duration of the simulation.

Note that the choice of $W$ is important as it acts as the resolution of the synchronicity between the network and physics simulators. A very small window will result in more tightly coupled simulations at the expense of more communication overhead and slower runtimes. Since the state of the channel as reported by the physics simulator in \texttt{NetworkUpdate} is fixed in the network simulator for the entire simulation window, larger choices of $W$ will introduce inaccuracies into the simulation results. This inherent tradeoff between simulation synchronism and communication overhead can be tuned by the user to suit the target application.

\begin{algorithm}[t]
    \caption{Synchronization protocol.}
    \begin{algorithmic}[1]
        \State \textbf{Initialize:} Set initial time $t=0$ and window size $W$.
        \State \textbf{Start simulation:} Send \\
        \quad \texttt{Update} with \texttt{BEGIN} and $\texttt{time\_val}=0$
        \While{there are events to run} 
        \State \textbf{Wait for synchronism:} Wait for\\
        \quad\quad \texttt{BEGIN} with $\texttt{time\_val}=t$
        \State \textbf{Report finished window:} Send\\
        \quad\quad \texttt{Update} with \texttt{END} and $\texttt{time\_val}=t$      
        \State \textbf{Update timestamp:}  \\
        \quad \quad $t=t + W$
        \State \textbf{Request next window:} Send \\        
        \quad\quad \texttt{Update} with \texttt{BEGIN} and $\texttt{time\_val}=t$      
        \EndWhile
    \end{algorithmic}
    \label{alg:sync_alg}        
\end{algorithm}

\section{Simulations}
\label{sec:numerical}

\begin{figure}[t]
    \centering
    \includegraphics[scale=1]{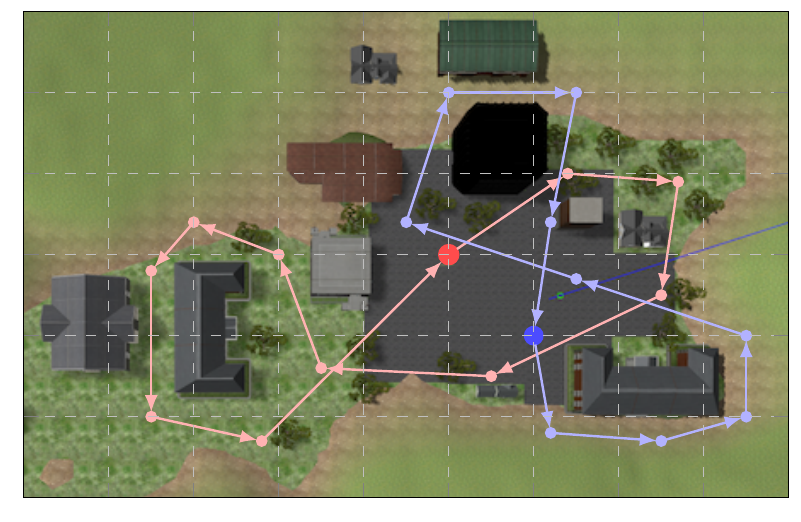}     
    \caption{Perimeter patrol scenario. Two agents perform a patrol task by following the trajectories outlined and starting from the two positions marked with a brighter point. The size of the environment is around $180 \times 120$ meters with each grid line corresponding to approximately $20$ meters.}
    \label{fig:test_scenario}
\end{figure}

To demonstrate the capabilities of ROS-NetSim, we have simulated the scenario illustrated in Fig. \ref{fig:test_scenario}. We consider a patrol task, in which two agents are moving along the pre-specified paths shown in Fig. \ref{fig:test_scenario}. The size of this environment is around $180 \times 120$ meters, with each square corresponding to approximately $20 \times 20$ meters. Several buildings of varying sizes and some minor vegetation populate the environment. As the agents move around, the propagation conditions of their wireless signal will be affected by these environmental factors.

Regarding our choice of robotic platform, we use UAVs as the patrolling robots. Specifically, we consider a simulation of the 3DR Iris quadrotor and use the PX4 autopilot flight stack, run as Software-In-The-Loop (SITL), together with Gazebo as our physics simulator. On the wireless communications side, we equip each of the agents with an IEEE 802.11 (Wi-Fi) interface, operating in ad-hoc mode. To simulate this equipment, we use a system level simulator, the WINTERSim network simulator \cite{wintersim}. The use of this system level simulator allow us to accurately model the communication process to simulate realistic conditions corresponding to those of an IEEE 802.11n stack. We consider unidirectional communication, in which the the red agent attempts to communicate as much data as possible with the blue agent over a TCP/IP data link. Tying Gazebo and WINTERSim together, we launch ROS-NetSim with a synchronization window of $1$ ms and a LOS/NLOS channel abstraction.

\begin{figure}[t]
	\centering
    \includegraphics[scale=1]{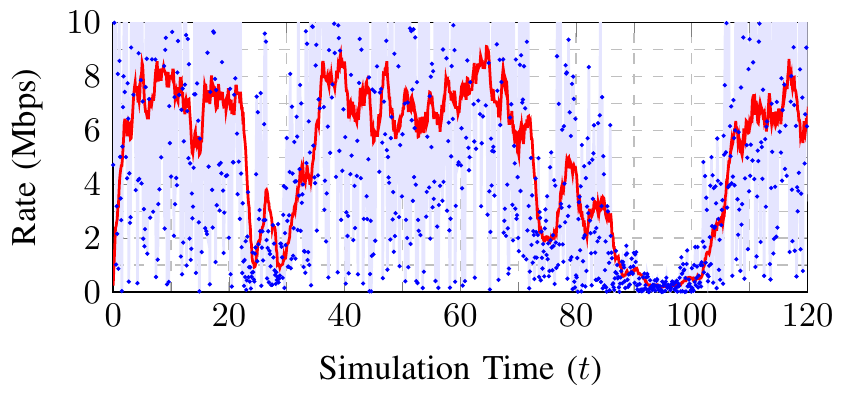}     	
	\caption{Average communication rate observed over the duration of the patrol task. Each blue point corresponds to a measurement with a spacing of $10$ ms. A moving average over a $0.2$ s window is shown in red.}
	\label{fig:rateTime}
\end{figure}

We simulate the system for the duration required for the two agents to do a single loop around their perimeter and obtain the following results. In Figure \ref{fig:rateTime} we show the achieved transmission rate as the simulation evolves over time. As the agents start relatively close to each other ($\sim30$ m apart) and without any obstacles between them (LOS communication), their attained rates hover around $8$ Mbps. As the agents take off and initiate their patrol, they take almost opposite directions, distancing themselves from each other. Specially, as the blue agent goes behind the building on the bottom right of the environment and the red agent goes around a small cluster of buildings, the transmission rate drops significantly to around $1-2$ Mbps (around the interval $20-30$ seconds in the simulation). This severe drop in transmission rate is caused by the non-line-of-sight propagation conditions brought by the environment. As both agents continue their patrol, they get close to each other and again regain line of sight, resulting in a quick recovery of the communication rate to around $8$ Mbps (during the interval $40-60$ seconds in the simulation). The worst communication conditions are experienced as the red agent reaches the building on the left side of the environment. This causes the two agents to be more than $100$ meters apart and under non-line-of-sight conditions, with several building occluding the communication path. Under these conditions, the agents are effectively out of range of each other and communication is practically lost, with the communication rate being almost zero (around $80-100$ seconds in the simulation). As the agents go back to their starting positions, they get close to each other and start regaining good communication rates.

\begin{figure}[t]
	\centering
    \includegraphics[scale=1]{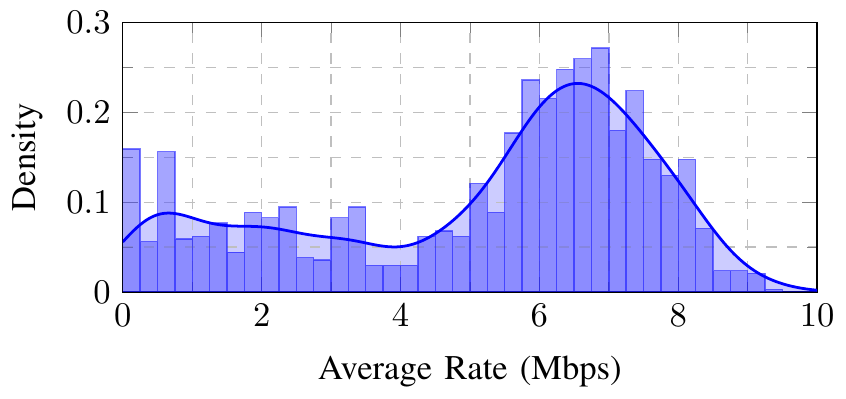}     
	\caption{Normalized histogram and an overlaid probability density estimate of the average communication rate experienced over the simulation of the patrol task.}
	\label{fig:rateHist}
\end{figure}

In order to grasp a better understanding of the communication rate experienced in this scenario, we plot in Fig. \ref{fig:rateHist} a normalized histogram and a probability density estimate of the average communication rate over a single simulated run. The average rate distribution allows us to see in a more detailed manner what we observed in the previous figure. Mainly, we observe that the communication rate hovers around $6-7$ Mbps for most of the task. More so, it more or less follows a Gaussian distribution centered around $7$ Mbps, with a long tail trailing to zero. We can infer from this figure that the agents spend most of their time under relatively good communication conditions. Furthermore, the time spent under bad conditions is nearly evenly distributed. Were the agents to spend more time in poor or NLOS conditions we would expect to see a second peak appear close to zero rate.

\begin{figure}[t]
	\centering
    \includegraphics[scale=1]{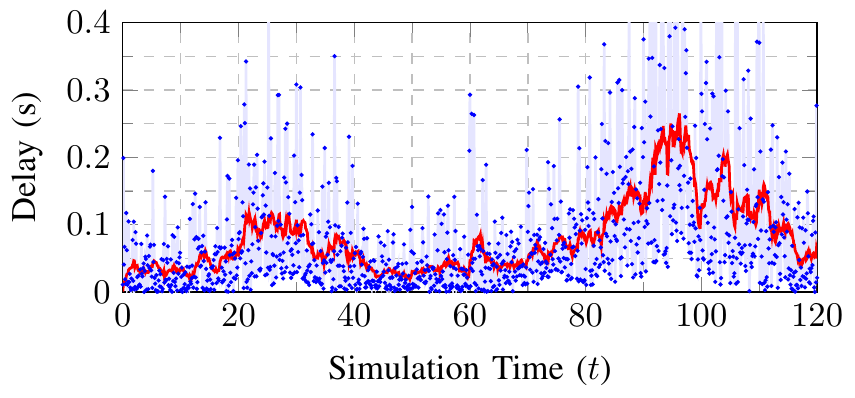}     	
	\caption{Average packet delay observed over the duration of the patrol task. Each blue point corresponds to a measurement with a spacing of $10$ ms. A moving average over a $0.2$ s window is shown in red.}
	\label{fig:delayTime}
\end{figure}

\begin{figure}[t]
	\centering
    \includegraphics[scale=1]{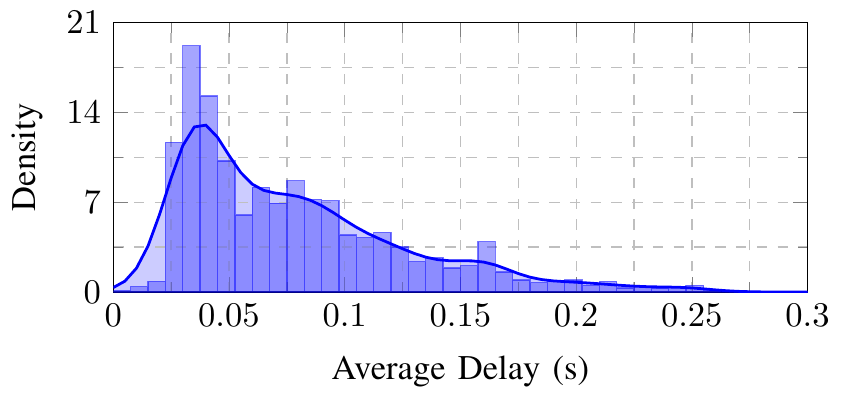}     
	\caption{Normalized histogram and an overlaid probability density estimate of the average packet delay experienced over the simulation of the patrol task.}
	\label{fig:delayHist}
\end{figure}

Now, we switch our attention to a different metric of interest: communication latency or packet delay. We show in Figure \ref{fig:delayTime} the packet delay as the simulation evolves over time, where we obtain delay measurements every $10$ ms and produce a moving average over a $0.2$ seconds window. It is clear from the figure that delay spikes during the same periods where the rate drops, during the $20-30$ and $80-100$ seconds intervals identified previously. Certainly, this is not surprising, as an increase in packet error rates will cause at the same time the communication rate to drop (due to effectively receiving less data) and the delay to increase (due to the increase in retransmissions). However, this effect is more pronounced in the communication rate.

We also plot in Figure \ref{fig:delayHist} the normalized histogram of average packet delay and a corresponding probability density estimate. From this characterization, we observe that the average delay distribution has a main mode around $40$ ms and a long exponential tail extending towards larger delays. In general, most of the events are covered by $0.25$ seconds of delay and delays under 25 ms are almost never experienced.

\begin{figure}[t]
	\centering
    \includegraphics[scale=1]{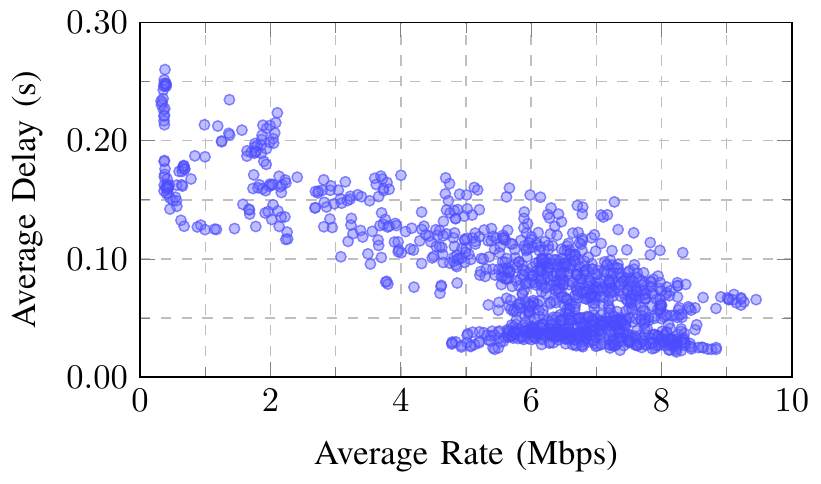}     	
	\caption{Scatter plot of the average transmission rate against the average packet delay experienced over the simulation.}
	\label{fig:scatter}
\end{figure}

Finally, we aim to study the relationship between transmission rates and delay. To this end, we obtain the scatter plot shown in Figure \ref{fig:scatter}. Here we have plotted the average data rate shown in Fig. \ref{fig:rateTime} against the average delay shown in Fig. \ref{fig:delayTime}. Two system behaviors can be inferred from this figure. First, most the time, the system is in the ranges previously identified. An average data rate of $6-8$ Mbps and a delay of $50-100$ ms. Second, and possibly more importantly, there is a clear link between rate and delay. This relationship can be summarized as follows: as the average delay increases (mainly caused by packet retransmissions)  the average communication rate decreases.

\section{Conclusions}
\label{sec:conclusions}

In this work we have introduced ROS-NetSim, a system designed to provide integration between robotic and network simulators. This platform has been designed with flexibility in mind, allowing the user to work with a wide array of physics and network simulations. The key component providing this flexibility is an interface between the simulators and a modular channel abstraction. This modularity coupled with ROS-NetSim's wide-ranging simulator compatibility, allows us to provide support for the scalable simulation of complex robotic and network behavior. Furthermore, we have aimed for transparent integration into ROS applications. As such, our system can be integrated into existing ROS projects with only configuration information needed to be specified. We have shown in numerical results, the ability to extract detailed wireless communication metrics from a simulation using our architecture.

\subsection{Extensions of the Proposed Framework}
\label{sec:extensions}

ROS-NetSim is intended to accurately simulate the communication traffic of a target ROS application in a single-master networked ROS setup. Sometimes, multi-master ROS systems are used by the community when performing decentralized experiments \cite{hernandez2015multi}, which would result in a mismatch between simulation and experiment. In this sense, extensions to the ROS-NetSim packet capture mechanism could be devised to increase the simulation accuracy of multi-master systems. Furthermore, while ROS-NetSim has been implemented as a ROS package, the design principles introduced in this work can be potentially extended to other systems. Of special interest is ROS2, which provides more suitable multi-agent support due to its use of Data Distribution Service (DDS) \cite{maruyama2016exploring}. The main functionalities of ROS-NetSim, i.e., physics and network coordination, modular channel geometry abstraction and its ability to capture traffic via the use of  virtual network interfaces are readily extendable to ROS2.

Furthermore, the ROS-NetSim platform can also potentially be used for hardware-in-the-loop simulation. Due to the transparency of ROS-NetSim with regard to the ROS target application, the ROS nodes themselves could be running on hardware, resulting in a hardware-in-the-loop simulation, with interactions with the physics and network simulators being handled by ROS-NetSim in a seamless manner.

\bibliographystyle{ieeetr}
\bibliography{references}

\end{document}